\documentclass[nohyperref]{article}

\usepackage{microtype}
\usepackage{graphicx}
\usepackage{siunitx}
\usepackage{subfigure}
\usepackage{booktabs} 

\usepackage{hyperref}

\usepackage[accepted]{icml2023}

\usepackage{amsmath}
\usepackage{amssymb}
\usepackage{mathtools}
\usepackage{amsthm}
\usepackage{enumitem}
\usepackage{tabularx}
\newcolumntype{C}{>{\centering\arraybackslash}X}
\usepackage{multirow} 
\usepackage[capitalize,noabbrev]{cleveref}

\theoremstyle{plain}

\theoremstyle{definition}

\theoremstyle{remark}

\usepackage[textsize=tiny]{todonotes}

\icmltitlerunning{Last-Layer Fairness Fine-tuning}

\begin{document}

\twocolumn[
\icmltitle{Last-Layer Fairness Fine-tuning is Simple and Effective for Neural Networks}



\icmlsetsymbol{equal}{*}

\begin{icmlauthorlist}
\icmlauthor{Yuzhen Mao}{sfu}
\icmlauthor{Zhun Deng}{harvard}
\icmlauthor{Huaxiu Yao}{stf}
\icmlauthor{Ting Ye}{uw}
\icmlauthor{Kenji Kawaguchi}{nus}
\icmlauthor{James Zou}{stf}
\end{icmlauthorlist}

\icmlaffiliation{sfu}{Simon Fraser University}
\icmlaffiliation{harvard}{Harvard University}
\icmlaffiliation{stf}{Stanford University}
\icmlaffiliation{uw}{University of Washington}
\icmlaffiliation{nus}{National University of Singapore}

\icmlcorrespondingauthor{Zhun Deng}{zhundeng@g.harvard.edu}
\icmlcorrespondingauthor{James Zou}{jamesz@stanford.edu}

\icmlkeywords{Machine Learning, ICML}

\vskip 0.3in
]



\printAffiliationsAndNotice{}  


\begin{abstract}
 As machine learning has been deployed ubiquitously across applications in modern data science, algorithmic fairness has become a great concern. Among them, imposing fairness constraints during learning, i.e. in-processing fair training, has been a popular type of training method because they don't require accessing sensitive attributes during test time in contrast to post-processing methods. While this has been extensively studied in classical machine learning models, their impact on deep neural networks remains unclear. Recent research has shown that adding fairness constraints to the objective function leads to severe over-fitting to fairness criteria in large models, and how to solve this challenge is an important open question. To tackle this, we leverage the wisdom and power of pre-training and fine-tuning and develop a simple but novel framework to train fair neural networks in an efficient and inexpensive way --- last-layer fine-tuning alone can effectively promote fairness in deep neural networks. This framework offers valuable insights into representation learning for training fair neural networks. The code is published at~\url{https://github.com/yuzhenmao/Fairness-Finetuning}
\end{abstract}

\vspace{-2em}
\section{Introduction} \label{sec:intro}
  The social impacts of machine learning systems deployed in our daily lives are getting increasing attention in modern data science. Great efforts have been put into understanding and correcting biases in algorithms \citep{hardt2016equality,dwork2012fairness}. However, most of the research has been conducted on understanding and correcting biases in classical machine learning models and simple datasets such as regression models and the adult income dataset \citep{ding2021retiring}. In contrast, in modern data science, tasks are more complex (e.g. classification on high dimensional datasets such as images), and over-parameterized models such as neural networks are deployed, which have been proven to reach the state-of-art in prediction performance. Thus, it is critical to study and understand how fairness techniques work on modern architectures. 

 
 Among all fairness techniques, in-processing fair training, which imposes fairness constraints during learning, have been a popular type of fair training method given the advantage of not requiring to access sensitive attributes during test time as post-processing methods \citep{kim2019multiaccuracy,hardt2016equality} and can more efficiently use the information of labels compared with pre-processing methods \citep{madras2018learning}. However, as pointed out by \citet{cherepanova2021technical} in-processing techniques are less effective for over-parameterized large neural networks because the model can easily overfit the fairness objectives during training, especially when the training data is imbalanced. \citet{cherepanova2021technical} raised this fairness over-fitting issue as an open challenge. Although  
 there has been work \cite{deng2022fifa} trying to address the challenge, the computational cost is much more expensive compared to standard training.
 

In this paper, we aim to tackle the challenge mentioned above and avoid the issue of overfitting the fairness criteria in an \textit{efficient and inexpensive} way when training neural networks. We focus on pre-training and fine-tuning, which have been proven to be useful techniques in obtaining powerful neural networks and are widely applied in state-of-the-art object detection. According to \citet{kirichenko2022last}, re-training the last layer of suitably pre-trained representations can reduce vulnerability to spurious correlation, and thus significantly improve prediction accuracy on imbalanced dataset and model robustness to covariate shift. A natural question induced by that is:
\begin{center}
``\textit{Will fine-tune the last layer or a small fraction of a standard-trained neural network with fair training methods be enough to obtain a fair neural network?}"
\end{center}

We provide a positive answer to the above question. As our \textit{main contribution}, we leverage the wisdom and power of pre-training and fine-tuning so as to develop a simple but effective framework to train fair neural networks in an efficient and inexpensive way (as illustrated in Figure 1 in the appendix):
(1) pre-training to obtain a representation by standard empirical risk minimization; (2) fine-tuning a few extra layers of neural networks by imposing fairness constraints while fixing the obtained representation (more details in Section~\ref{sec:fdfr}).  In addition, we further show that our method can even work for out-of-domain data by fine-tuning while we only train the representation on a source dataset. Finally, we also explore whether fine-tuning other structures beyond the last layer of neural networks can perform well (see details in Section~\ref{sec:surg}).

\section{Preliminaries}\label{sec:pre}
\subsection{Notation}
We consider a dataset consists of triplets, i.e. $\mathcal{D}=\left\{\left(x_i, a_i, y_i\right)\right\}_{i=1}^N$, where for each triplet, $x_i$ is the feature vector drawn from an input distribution over $\mathcal{X}$, $a_i \in \mathcal{A}$ denotes the corresponding sensitive attribute such as race or gender, and $y_i \in \mathcal{Y}$ is the corresponding label. Throughout the paper, for simplicity, we only consider $a$ and $y$ as binary variables, where $a \in \{0, 1\}$ and $y \in \{0, 1\}$. However, our method can easily be generalized to multiple sensitive attributes and multi-class scenarios. We further denote the cross entropy loss as $\hat{L}(h)=- \sum_{i=1}^m y_i \cdot \log \left(p(h(x_i))\right)/m$, where $p\circ h$ is the estimation of the prediction probability for the correct class for a sample $x_i$, and $p$ is a soft-max function.

\subsection{Fairness notions}

In standard supervised learning tasks, people aim to train a model $h \in \mathcal{H}: \mathcal{X} \mapsto \mathcal{Y}$, where $\hat{y_i} = h(x_i)$ is the prediction of $y_i$ for a given feature vector $x_i$.  For example, in the CelebA dataset~\citep{liu2015faceattributes}, the task is to classify the hair color of the celebrity in the image and use gender as the sensitive attribute. In CelebA dataset, there are four different groups corresponding to four different combinations of ($y_i$, $a_i$): Blonde woman ($\mathcal{G}_1$), blonde man ($\mathcal{G}_2$), Non-blonde woman ($\mathcal{G}_3$) and Non-blonde man ($\mathcal{G}_4$). Since $\mathcal{G}_2$ contains only 1\% images of the whole dataset, it is referred as the \textit{minority group}. For larger groups such as $\mathcal{G}_1$ and $\mathcal{G}_4$, we refer them as \textit{majority groups}. This imbalance existing in the dataset usually results in an unfairly biased model with standard training. To de-bias the model and make it ``fairer", people propose adding varieties of additional fairness constraints (regularization terms) to objective functions to achieve model fairness. Specifically, in this paper, we mainly focus on the following three popular fairness notions which have been widely used in the previous literature: Equalized Odds (EO), Accuracy Equality (AE) and Max-Min Fairness (MMF). In this paper, we don't discuss notions such as  Disparate Impact or Demographic Parity because their definitions are problematic in the way that they cannot distinguish qualified individuals from others in each group \citep{hardt2016equality} and in general cannot align with the model accuracy well.

\textbf{\textbf{Equalized odds (EO)}.}  Equalized odds requires, given the true label $y$, an algorithm's decisions/outcomes do not depend on the sensitive attributes of individuals, such as race, gender, or age, which indicates that $\hat{y}$ is conditionally independent of the sensitive attribute $a$ given $y$. In other words, it means that the false positive rate and false negative rate should be the same for all groups, so that no group is unfairly disadvantaged. Equalized odds~\citep{hardt2016equality} is defined as:
\begin{equation}
\mathbb{P}(\hat{y}=1 \mid a=0, y=y) = \mathbb{P}(\hat{y}=1 \mid a=1, y=y)
\end{equation}

To enforce equalized odds in practice, one way people implement in training is to minimize the following objectives~\citep{manisha2018fnnc, cherepanova2021technical}:
\begin{equation}
\min _h\left[\hat{L}_{w}\left(h\right)+\alpha(f p r+f n r)\right]
\label{eq:2}
\end{equation}
for a given predefined weight $\alpha$, where
$$
\begin{aligned}
& f p r=\left|\frac{\sum_i p_i\left(1-y_i\right) a_i}{\sum_i a_i}-\frac{\sum_i p_i\left(1-y_i\right)\left(1-a_i\right)}{\sum_i\left(1-a_i\right)}\right| \\
& f n r=\left|\frac{\sum_i\left(1-p_i\right) y_i a_i}{\sum_i a_i}-\frac{\sum_i\left(1-p_i\right) y_i\left(1-a_i\right)}{\sum_i\left(1-a_i\right)}\right| .
\end{aligned}
$$
Here, $p_i$ denotes a softmax output (binary prediction task) of the model $h$ and $\hat L_{w}$ is the weighted cross-entropy loss.



\textbf{\textbf{Accuracy equality} (AE).} Accuracy equality requires an algorithm produces outcomes that are (approximately) equally accurate for individuals belonging to different protected groups. Its goal is to ensure that an algorithm does not unfairly advantage or disadvantage certain groups, and instead provides equally accurate predictions for all individuals. In other words, a model satisfies accuracy equality if its misclassification rates are equal across different sensitive groups~\citep{zafar2017fairnessb}. Accuracy equality is defined as:
\begin{equation}
\mathbb{P}(\hat{y} \neq y \mid a=0) = \mathbb{P}(\hat{y} \neq y \mid a=1)
\end{equation}
To enforce accuracy equality in practice, one way people implement in training is to minimize the following objective:
\begin{equation}
\min _h\left[\hat{L}_{w}\left(h\right)+\alpha\left|\hat{L}^{a+}\left(h\right)-\hat{L}^{a-}\left(h\right)\right|\right]
\label{eq:4}
\end{equation}
 for a given predefined weight $\alpha$, where $\hat{L}^{a+}(h)$ is the cross entropy loss of samples with $a = 1$, and $\hat{L}^{a-}(h)$ is the cross entropy loss of samples with $a = 0$. $\hat L_{w}$ is the weighted cross-entropy loss.

\textbf{\textbf{Max-Min fairness} (MMF).} Max-min fairness focuses on maximizing the performance of the worse-off group,
i.e., the group with the lowest utility~\citep{lahoti2020fairness,cherepanova2021technical}. It is defined as:
\begin{equation}
\max \min _{y \in  \mathcal{Y}, a \in \mathcal{A}} \mathbb{P}(\hat{y}=y \mid y, a)
\end{equation}

To satisfy max-min fairness, people aim to minimize the following objective \citep{cherepanova2021technical}:
\begin{equation}
\begin{aligned}
\min _h \max 
& \{\hat{L}^{(y+,a+)}\left(h\right), \hat{L}^{(y+,a-)}\left(h\right), \\
& \hat{L}^{(y-,a+)}\left(h\right), \hat{L}^{(y-,a-)}\left(h\right)\},
\end{aligned}
\label{eq:6}
\end{equation}
where $\hat{L}^{(y',a')}\left(h\right)$ denotes the cross-entropy loss on the training samples where $y = y'$ and  $a = a'$.

\vspace{-1em}
\section{Problem Background}\label{sec:problem}
\subsection{Challenges in training fair neural networks with in-processing techniques}\label{subsec:cha}
 To tackle fairness concerns in modern data science, researchers have proposed and formalized various notions of fairness as well
as methods for mitigating unfair behavior. However, as pointed out by \citet{cherepanova2021technical}, \citet{deng2022fifa} and our introduction section, the effectiveness of in-processing techniques in fairness that impose fairness constraints on modern structures such as deep neural networks
is unclear and still under exploration. Specifically, \citet{cherepanova2021technical} observe that
large models overfit to fairness objectives and
produce a range of unintended and undesirable
consequences by conducting studies on
both facial recognition and automated medical
diagnosis datasets using state-of-the-art architectures.
They empirically emphasize the over-fitting issue in training fair neural networks, where models trained with fairness constraints become too closely aligned with the training data, leading to poor performance on unseen data in terms of fairness.

Their studies are mainly based on two main approaches for rectifying unfair behavior. (1) The first one is to impose the fairness constraints or regularizers on the training objective and train the \textit{full} neural network. Based on the experiments, they find the model shows excellent performance on the training set and appears to be fair, in terms of the difference in AUC (see details in Section~\ref{sec:exp}) values for different sensitive attributes. However, upon evaluation on the test set, models trained with fairness constraints can be proven to be even less fair compared to a baseline model, which indicates a very serious over-fitting issue. Increasing the strength of the constraints results in a higher accuracy trade-off, but it still fails to significantly improve fairness on the validation and test sets. Cherepanova et al. attribute this to the over-parameterized nature of deep neural networks and the fluid decision boundary it creates. (2) They also try to only apply fairness penalties on a holdout set after training a model without fairness constraints and fine-tune the \textit{full} neural network on the hold-out set. They assume that the ineffectiveness of fairness constraints on the training set is due to the high training accuracy, which makes the models appear fair regardless of their performance during testing. However, the issue of over-fitting remains prevalent, that is, the fairness is achieved on the train dataset but cannot be generalized to unseen data.


\subsection{Our Inspiration: standard training can still learn core features on imbalanced datasets}\label{sec:core}
In addition to the previous success of pre-training and fine-tuning in mitigating the issue of spurious correlation \citep{kirichenko2022last}, our inspiration is also drawn from the observation that core features can still be learned by standard training on imbalanced datasets, which can be used to enable accurate predictions for minority groups in the later fine-tuning phase. 

To access this property, we evaluate the ResNet-18 model that has been trained on the original CelebA dataset using empirical risk minimization (ERM), on a customized hair-only dataset $D_{H}$, which contains 29,300 sampled images of the hair segmented from the original training images using the mask from~\citet{CelebAMask-HQ} on uniform grey background. In order to effectively examine whether the model has learned the core features that are relevant to the labels and whether these features have been properly encoded in the preceding layers before the final layer, we divide the $D_H$ into two sets: $D_H^{Tr}$ and $D_H^{Te}$. $D_H^{Tr}$ is evenly balanced and comprises 107 images from each $(a,y)$ group, totaling 428 images. On the other hand, $D_H^{Te}$ includes the remaining 28,872 images, with at least 107 images per group. Then we fine-tune the last layer of the model on $D_H^{Tr}$ and evaluate it on $D_H^{Te}$. We repeat this process with the model pre-trained on a balanced dataset sampled from the original training dataset, as a comparison to the model pre-trained on the original imbalanced dataset. We present the mean and the worst group accuracy of all the experiments mentioned above in Table~\ref{tab:s5}. Additionally, we also include the results on the original test dataset without fine-tuning for comparison purposes. Based on the results, we find that there is a significant discrepancy between the mean and worst group accuracy on the original test data, which means that the standard-trained full model is heavily influenced by the imbalanced training dataset, leading to poor performance on minority groups. On the other hand, we find the last-layer-fine-tuned model achieves very good performance on the hair-only dataset with 86\% and 82\% worst group accuracy. This results indicate that although the model trained on the imbalanced dataset under-performs on the minority groups, it can still learn the core features which are relevant to the classification task, and only need simple fine-tuning to perform accurate predictions on the core features. This conclusion also well supports the method we analyze in the following sections. 
\begin{table}[ht]
\centering
\setlength{\tabcolsep}{3pt}
\begin{tabular}{c|c|c}
\hline
 {\multirow{2}{*}{Train} } & \multicolumn{2}{c}{ Test (Worst/Mean) } \\ \cline { 2 - 3 }
 
 & Original & Hair-only \\ \hline
  Original & 0.268/0.946 & 0.863/0.878   \\ \hline
  Balanced & 0.789/0.835 & 0.827/0.843   \\ \hline
\end{tabular}
\caption{Representation learning on CelebA. The column Original corresponds to directly evaluate on the original test dataset. The column Hair-only corresponds to last-layer fine-tuned results on the sampled hair-only dataset $D_H^{Te}$.}
\label{tab:s5}
\end{table}

\vspace{-1em}
\section{Our Main Approach: Fair Deep Feature Reweighting}\label{sec:fdfr}

In \citet{cherepanova2021technical}, the authors point out over-parameterization of a neural network lead to the over-fitting of fairness criteria since over-parameterization makes the learned neural network's decision boundary highly flexible, and trying to fit the boundaries to meet fairness criteria for one attribute can negatively impact fairness with regards to another sensitive attribute. However,over-parameterization has been a key for neural networks to achieve high accuracy in prediction, especially for those neural networks designed to tackle challenging tasks. Inspired by previous work \citep{kirichenko2022last} on spurious correlation, in this section, we show how to solve this dilemma in a simple way based on pre-training and fine-tuning, and we call our method \textit{fair deep feature reweighting}. Our method not only provides answers to the challenges brought up by \citet{cherepanova2021technical}, but it is also surprisingly simple and computationally cheap. Our approach also reveals an interesting future direction for research on fairness.

\textbf{\textbf{Step 1: pre-train a representation.}} As discussed in Section~\ref{sec:core} and previous works~\citep{lee2022surgical, kirichenko2022last}, standard training by doing empirical risk minimization (ERM) is enough to obtain representations that capture core features of input in many cases. With this spirit, we first train a neural network $\mathcal N$ with ERM and obtain a representation $\Phi$, which is the sub-neural-network from the first layer to the next to the last layer of $\mathcal N$, i.e. $\mathcal N=w\circ \Phi$, where $w$ is the last layer.

\textbf{\textbf{Step 2: fine-tune the last layer with reweighting and fairness constraints.}} We then fix $\Phi$ and improve model fairness through last-layer fine-tuning by incorporating fairness constraints (Eq.~\ref{eq:2} \&~\ref{eq:4} \&~\ref{eq:6}) and data reweighting obtain a new last layer $w^{new}$ and another neural network $\mathcal N^{new}=w^{new}\circ \Phi$. Specifically, for data reweighting, we first sample a small dataset $D_r$ from the training dataset $D$ and the validation dataset $\hat{D}$. Each $(a, y)$ group in $D_r$ has the same number of samples, where $a$ and $y$ represent the sensitive attribute and the label respectively. We then train $w^{new}$ from scratch on the balanced dataset $D_r$ with standard ERM and fairness constraints.

In summary, fair deep feature reweighting,  which only requires updating the parameters of the last layer when training with fairness constraints, avoids the over-parameterization issue described in~\citet{cherepanova2021technical} and reduces the risk of over-fitting. In addition, with respect to fairness, data reweighting can also be particularly beneficial in correcting models that have been negatively impacted by imbalanced training datasets, where one class is heavily overrepresented compared to others. Our approach allows the model to better capture patterns in the data, leading to improved fairness in its performance.

\textbf{\textbf{Intuition behind our approach.}} As implied by our observation in Section~\ref{sec:core}, ERM can already encode the information of core features well in a representation and we only need to further compose the representation with a linear structure to recover those core features and used them to predict. Since the fairness criteria are imposed only at the fine-tuning phase, we will not suffer from over-fitting issue.

\vspace{-1em}
\section{Experiments} \label{sec:exp}

\subsection{Experimental Setup}
In this section, we briefly discuss our experimental setup and put detailed descriptions in Appendix~\ref{sec:experimental_setup_app}.

\textbf{Datasets and Hyperparameters.}
To evaluate the performance of our methods, we conduct a comparative analysis using various baselines on the CelebA and UTKFace datasets, which focus on facial recognition. Our model architecture is based on ResNet-18, as employed in previous studies \citep{cherepanova2021technical}. During the fine-tuning process, we replace the last layer with a newly initialized layer and subsequently update only this layer while keeping the remaining model parameters fixed. To train our model, we utilize SGD with a momentum of 0.9 and weight decay of 5e-4. All hyperparameters are carefully tuned using a separate validation dataset, and we provide a summary of these hyperparameters in Table~\ref{tab:hyper}. We evaluate the model's fairness and prediction.

\textbf{Compared methods.}~\label{bsl} We define baseline methods and our proposed method FDR as follows:



\noindent\underline{FullFT-Reg}: Impose
the fairness constraints on the training objective and train the full neural network.

\noindent \underline{LastFT}: Fine-tune the last layer of a pre-trained model on the imbalanced validation dataset~\citep{kirichenko2022last}.

\noindent\underline{LastFT-RW}: Fine-tune the last layer of a pre-trained model on the balanced dataset.

\noindent\underline{LastFT-Reg}: Fine-tune the last layer of a pre-trained model on the validation dataset with fairness constraints.

\noindent \underline{FDR (\textbf{ours})}: Fine-tune the last layer of a pre-trained model on the balanced dataset with fairness constraint.


\textbf{Metrics.} To assess the accuracy of our model, we employ weighted accuracy (WACC) and Area under the ROC Curve (AUC) as evaluation metrics, as suggested by~\citet{fawcett2004roc}. For evaluating fairness, we employ the metrics Equalized Odds Difference (EO\_Diff), Accuracy Equality Difference (AE\_Diff), and Worst Accuracy (WA). Additionally, we introduce a novel metric, denoted as AF, that takes into account both accuracy and fairness. Please refer to Appendix~\ref{sec:metric} for detailed descriptions of metrics.
\vspace{-0.5em}
\subsection{Assessing Fairness on CelebA Dataset}
\vspace{-0.2em}
We conducted an evaluation of various methods on the CelebA dataset (see the dataset details in Appendix~\ref{dataset_des}) with the goal of accurately predicting the hair color in each image. The results of three different fairness notions are presented in Table~\ref{tab:1}. Our experiments revealed the following findings:
\vspace{-1em}
\begin{itemize}[leftmargin=*]
\item Fine-tuning only the last layer of a pre-trained model, as opposed to fine-tuning all layers (FullFT-Reg), led to significant improvements in all fairness metrics and reduced their generalization gap between the training and test datasets. This indicates that last layer fine-tuning effectively addresses the issue of overfitting.
\vspace{-0.5em}
\item Last-layer methods that incorporated fairness constraints, i.e., LastFT-Reg and FDR, exhibited relatively lower test WACC compared to methods without such constraints. This suggests that adding fairness constraints may negatively impact the model's prediction accuracy.
\vspace{-0.5em}
\item Among the evaluated methods, FDR demonstrated the best performance in both fairness metrics and the accuracy-fairness (AF) evaluation. This implies that last layer fine-tuning, combined with data reweighting and fairness constraints, can efficiently and effectively mitigate the overfitting issue.
\end{itemize}
\vspace{-1em}
Apart from the above results, we conduct experiments under transfer learning setting and using surgical fine-tuning~\cite{lee2022surgical} in Appendix~\ref{trans} and~\ref{sec:surg}, respectively. All results support our claims.

\begin{table}[t]
\caption{Overall performance on CelebA dataset with different fairness constraints, where averaged values are reported over twenty random trials.}
\vspace{0.2em}
\centering
\resizebox{\linewidth}{!}{\setlength{\tabcolsep}{1mm}{
\begin{tabular}{l|cc|cc|cc|c}
\toprule
\textbf{Fairness Notion 1: EO} & \multicolumn{2}{c|}{WACC} & \multicolumn{2}{c|}{AUC} & \multicolumn{2}{c|}{EO\_Diff} & AF \\
\midrule
& Train & Test & Train & Test & Train & Test & Test \\
\midrule
FullFT-Reg & 1.000 & \textbf{0.914} & 1.000 & 0.969 & 0.000 & 0.499  & 0.415 \\
\midrule
LastFT & 0.918 & 0.913 & 0.974 & \textbf{0.971} & 0.308 & 0.327  & 0.586 \\
LastFT-RW & 0.913 & 0.908 & 0.970 & 0.968 & 0.100 & 0.207 & 0.701\\
LastFT-Reg & 0.898 & 0.901 & 0.971 & 0.969 & 0.177 & 0.153 & 0.748 \\
FDR & 0.898 & 0.892  & 0.962 &  0.958 & 0.031 &  \textbf{0.107} & \textbf{0.785} \\
\midrule
\midrule
\textbf{Fairness Notion 2: AE} & \multicolumn{2}{c|}{WACC} & \multicolumn{2}{c|}{AUC} & \multicolumn{2}{c|}{AE\_Diff} & AF \\
\midrule
& Train & Test & Train & Test & Train & Test & Test \\
\midrule
FullFT-Reg & 1.000 & \textbf{0.914} & 1.000 & 0.968 & 0.000 & 0.049 & 0.865 \\
\midrule
LastFT & 0.918 & 0.913 & 0.974 & \textbf{0.971} & 0.066 & 0.043 & 0.869 \\
LastFT-RW & 0.913 & 0.908 & 0.970 & 0.968 & 0.026 & 0.020 & 0.888\\
LastFT-Reg &0.907  & 0.904 & 0.969 & 0.964 & 0.016 & 0.009 & 0.895 \\
FDR & 0.898 & 0.900 & 0.963 & 0.967 & 0.009 & \textbf{0.003} & \textbf{0.897} \\
\midrule
\midrule
\textbf{Fairness Notion 3: MMF} & \multicolumn{2}{c|}{WACC} & \multicolumn{2}{c|}{AUC} & \multicolumn{2}{c|}{WA} & AF \\
\midrule
& Train & Test & Train & Test & Train & Test & Test \\
\midrule
FullFT-Reg & 1.000 & 0.910 & 1.000 & 0.969 & 1.000 & 0.393 &  1.303 \\
\midrule
LastFT & 0.918 & \textbf{0.913} & 0.974 & \textbf{0.971} & 0.633 & 0.598  & 1.511 \\
LastFT-RW & 0.913 & 0.908 & 0.970 & 0.968 & 0.872 & 0.732  & 1.640 \\
LastFT-Reg & 0.896 & 0.888 & 0.960 & 0.955 & 0.879
 & 0.717  & 1.605 \\
FDR & 0.902 & 0.898 & 0.964 &  0.962& 0.868
 & \textbf{0.803}  & \textbf{1.701} \\
\bottomrule
\end{tabular}}}
\vspace{-1em}
\label{tab:1}
\end{table}

\vspace{-1em}
\section{Conclusion}
\vspace{-0.3em}
In this paper, we present a simple yet innovative framework for training fair neural networks through the use of pre-training and fine-tuning. The experimental findings compellingly illustrate that fine-tuning the last layer alone with data reweighting is sufficient for promoting fairness in deep neural networks.


\bibliography{example_paper}
\bibliographystyle{icml2023}


\newpage
\appendix
\onecolumn
\begin{figure*}
\begin{center}
  \includegraphics[width=1\textwidth]{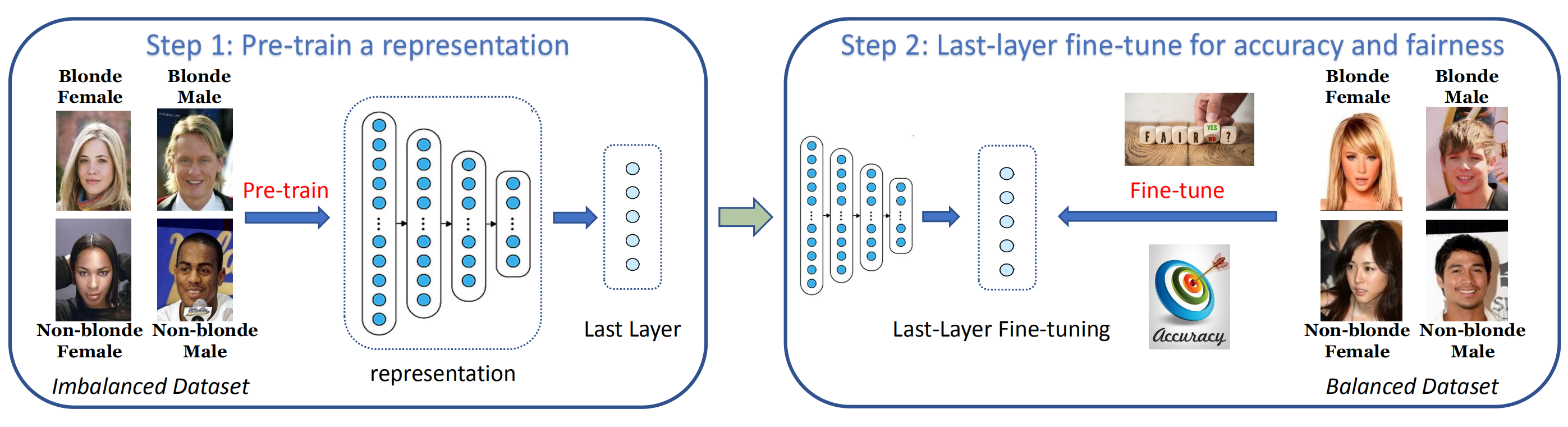}
  \caption{\textbf{Scheme of our approach.} We propose a simple framework to obtain fair neural networks by in-processing techniques. (1) We first obtain a representation by training a model via standard empirical risk minimization on the dataset (possibly imbalanced). (2) Next, we only fine-tune the last layer of the model on a balanced version of the original dataset via varities of in-processing techniques in fairness.}
\end{center}
  \label{fig:teaser2}
\end{figure*}


\section{Additional Related work}
\label{sec:related_work}
  \noindent\textbf{Fairness in machine learning.} 
Fairness in machine learning has been the focus of much research in recent years. A growing body of literature has aimed at addressing the potential biases in machine learning models with respect to sensitive attributes such as race, gender, and age~\cite{zafar2017fairnessb, kamiran2012data, hardt2016equality}. Researchers have proposed various fairness metrics and algorithms to mitigate such biases, including group fairness~\cite{dwork2012fairness,binns2020apparent,deng2022reinforcement}, individual fairness~\cite{dwork2011firm,petersen2021post}, and causal fairness~\cite{gerstenmayer2018analysis}. Another existing approach involves modifying the learning algorithm to ensure that the machine learning model is fair with respect to certain protected groups~\cite{zafar2017fairnessb}. This has led to the development of various algorithmic fairness techniques, such as data reweighting, adversarial debiasing, and regularization-based methods. The challenge lies in balancing fairness and accuracy, as the former can sometimes come at the cost of the latter~\cite{chouldechova2017fair}.

\noindent\textbf{Last layer fine-tuning.} Fixed-feature learning \cite{deng2020representation,deng2021adversarial,ji2021unconstrained,deng2021improving,burhanpurkar2021scaffolding,ji2021power,kawaguchi2022understanding} has been popular for its efficiency to adopt to out-of-domain data. Fine-tuning only a small fraction of the parameters can also effectively avoid over-fitting and bad generalization \cite{deng2021toward,kawaguchi2022robustness,zhang2020does,kawaguchi2023does}. Previous works~\cite{lee2022surgical, kirichenko2022last} have shown that instead of updating all the parameters of the model, only fine-tuning the last layer(s) can still match or even achieve better performance for spurious correlation and distribution shifts. Specifically, this approach involves training the last layer of a pre-trained neural network on a smaller, more specific dataset, with the aim of fine-tuning the model to perform well on this specific task. By using a pre-trained network as a starting point, last layer fine-tuning reduces the risk of over-fitting, as the initial layers have already learned high-level features from a large dataset. Another advantage of last-layer fine-tuning is that it can be time and computational resources efficient compared with training a network from scratch, since the pre-trained weights provide a good initialization point, allowing the network to converge faster. 

  
\section{Experiment Setup}
\label{sec:experimental_setup_app}
\subsection{Dataset Description}\label{dataset_des}
We mainly conduct our experiments on two popular datasets in facial recognition: CelebA~\citep{liu2015faceattributes} and UTKFace~\citep{zhang2017age}. Due to the page limit, we only show the results of CelebA in the main text, and present the results of UTKFace in Section~\ref{sec:exp
_add}. Both datasets are of high dimension and have been widely used in prior deep learning and fairness studies. We provide a comprehensive description of the datasets and the corresponding tasks as follows. We create the following four subsets for each dataset -- training set, validation set, test set, and an additional balanced dataset. 

\begin{itemize} [leftmargin=0.3cm]
     \item \textbf{CelebA}: a large-scale dataset of celebrity faces, including over 200k images with annotations of 40 different attributes such as facial landmarks, gender, age, hair color, glasses, etc~\citep{liu2015faceattributes}. In the experiment, we select hair color (blonde or not) as the label $y$ to predict, and use gender (male or not) as the sensitive attribute $a$. Specifically, (male, blonde hair) which only contains 1\% of the total images,  is the minority group of this dataset. We follow the dataset splitting format in the orignal paper and list the details in Tabel~\ref{tab:datasets1}. We construct a balanced sub-dataset by sampling from the original training and validation datasets based on the number of images in the minority group. Specifically, we select 1,569 images per ($y$, $a$) group, yielding a total of 6,276 images in the balanced dataset.
    \item \textbf{UTKFace}: a large, publicly available face dataset with long age span (range from 0 to 116 years old)~\citep{zhang2017age}. In our experiment, we randomly select 20\% data (maintain the same proportion) from the training dataset to serve as the validation dataset and randomly select another 20\% data (maintain the same proportion) from the training dataset to serve as the test dataset. The details of UTKFace dataset is presented in Tabel~\ref{tab:datasets2}. We select gender as the sensitive attribute and age as the label to predict. Following the setting of prior work~\citep{park2020readme}, the age feature is annotated into young ($\leq$ 35) and the others ($>$ 35). In UTKFace, we also create a balanced sub-dataset by sampling an equal number of images from the original training and validation datasets for each ($y$, $a$) group. Specifically, we select 2,477 images per group, resulting in a total of 9,908 images in the balanced dataset.

\end{itemize}
\begin{table}[hbt]
\caption{CelebA dataset statistics.}
\centering
\setlength{\tabcolsep}{4pt}
\begin{tabular}{c|c|c|c}
\toprule
  {(train/val/test)} & Blonde Hair & Non-blonde Hair   & Total \\ \midrule

  Male & 1,387/182/180 & 66,874/8,276/7,535 & 68,261/8,458/7,715   \\ \midrule
  Female & 22,880/2,874/2,480 & 71,629/8,535/9,767 & 94,509/11,409/12,247 \\   \midrule
  Total  & 24,267/3,056/2,660 & 138,503/16,811/17,302 & 162,770/19,867/19,962 \\  \bottomrule
\end{tabular}
\label{tab:datasets1}
\end{table}
\begin{table*}[hbt]
\caption{UTKFace dataset statistics.} 
\centering
\setlength{\tabcolsep}{4pt}
\begin{tabular}{c|c|c|c}
\toprule
  {(train/val/test)} & Young (age $\leq$ 35) & Old (age $>$ 35)   & Total \\ \midrule

  Male & 4,133/1,378/1,378 & 3,301/1,101/1,100 & 7,434/2,479/2,478   \\ \midrule
  Female & 4,931/1,643/1,644 & 1,858/619/619 & 6,789/2,262/2,263 \\   \midrule
  Total  & 9,064/3,021/3,022 & 5,159/1,720/1,719 & 14,223/4,741/4,741 \\  \bottomrule
\end{tabular}

\label{tab:datasets2}
\end{table*}

\subsection{Metrics}\label{sec:metric}
Besides the weighted accuracy ({WACC}) and Area under the ROC Curve ({AUC})~\citep{fawcett2004roc}, in terms of different fairness constraints tested in the experiment, we also use the following metrics to evaluate the model performance:
\begin{itemize} [leftmargin=0.3cm]   
    \item \underline{Equalized Odds Difference (the smaller the better):} 
    
    $\max \{|\mathbb{P}(\hat{y}=1 \mid a=0, y=0) - \mathbb{P}(\hat{y}=1 \mid a=1,
    y=0)|, |\mathbb{P}(\hat{y}=1 \mid a=0, y=1) - \mathbb{P}(\hat{y}=1 \mid a=1, y=1)|\}$



    \item \underline{Accuracy Equality Difference (the smaller the better):}

    $
    |\mathbb{P}(\hat{y} \neq y \mid a=0) - \mathbb{P}(\hat{y} \neq y \mid a=1)|
    $

    \item \underline{Worst Accuracy (the larger the better):}

    $\min \{ \mathbb{P}(\hat{y}=0 \mid a=0, y=0), \mathbb{P}(\hat{y}=0 \mid a=1, y=0), \mathbb{P}(\hat{y}=1 \mid a=0, y=1), \mathbb{P}(\hat{y}=1 \mid a=1, y=1) \}$ 


    
    
\end{itemize}
When report the experiment results, we use EO\_Diff, AE\_Diff, and WA to denote these metrics respectively. Furthermore, for the final evaluation criterion, we place equal weight on model prediction accuracy and fairness. To be more formal, we introduce an additional metric that combines both aspects - the \textbf{AF} metric. This metric is defined as an equal-weight linear combination of weighted accuracy and fairness metric, with both receiving equal weighting. Specifically, for Equalized Odds, AF = WACC - EO\_Diff; for Accuracy Equality, AF = WACC - AE\_Diff; while for Max-Min Fairness, AF = WACC + WA. We use different signs for different fair metrics, so that a larger AF value always indicates better performance of the model. We run all the experiments with twenty random seeds and report the mean over different trials.

\subsection{Hyper-parameters} 
The hyper-parameters shared among all methods are tuned using the search ranges shown in the Table~\ref{tab:hyper}. For all the four last layer methods, we use full-batch SGD to train the model. We select the values of the hyper-parameters that lead to the highest AF value for each method, and list them in Table~\ref{tab:hyper-celeba} and Table~\ref{tab:hyper-utk} for CelebA and UTKFace, respectively. 
\begin{table}[htp]
\caption{Hyper-parameter search ranges.}
\centering
\scalebox{1} {
\setlength{\tabcolsep}{4pt}
\begin{tabular}{c|c}
\toprule {\textbf{Hyper-parameter}}  &  \textbf{ search range}   \\
\midrule
{learning rate} & [\num{3e-4}, \num{1e-3}, \num{3e-3}] \\
{batch size} & Full \\
{number of epochs} & [500, 1000, 1500, 2000]  \\
{$\alpha$ (in Eq.~\ref{eq:2}\&\ref{eq:4})} & [0.5, 1, 2, 5, 10]  \\
\bottomrule
\end{tabular}
}
\label{tab:hyper}
\end{table}

\begin{table}[htp]
\caption{Hyper-parameter settings for CelebA.}
\centering
\setlength{\tabcolsep}{4pt}
\begin{tabular}{l|c|c|c}
\toprule
\textbf{Methods} & learning rate & number of epochs & $\alpha$ \\
\midrule
{LastFT} & \num{1e-3} & 1000 & /  \\
{LastFT-RW} & \num{3e-3} & 1000 & / \\
{LastFT-Reg (EO)} & \num{1e-3} & 500 & 10  \\
{FDR (EO)} & \num{1e-3} & 1000 & 2 \\
{LastFT-Reg (AE)} & \num{1e-3} & 1000 & 0.5  \\
{FDR (AE)} & \num{1e-3} & 500 & 5 \\
{LastFT-Reg (MMF)} & \num{1e-3} & 1000 & /  \\
FDR (MMF) & \num{1e-3} & 1000 & / \\
\bottomrule
\end{tabular}
\label{tab:hyper-celeba}
\end{table}

\begin{table}[htp]
\caption{Hyper-parameter settings for UTKFace.}
\centering
\setlength{\tabcolsep}{4pt}
\begin{tabular}{l|c|c|c}
\toprule
\textbf{Methods} & learning rate & number of epochs & $\alpha$ \\
\midrule
{LastFT} & \num{1e-3} & 1000 & /  \\
{LastFT-RW} & \num{3e-3} & 1000 & / \\
{LastFT-Reg (EO)} & \num{1e-3} & 1000 & 0.5  \\
{FDR (EO)} & \num{1e-3} & 1500 & 2 \\
{LastFT-Reg (AE)} & \num{1e-3} & 1000 & 1  \\
{FDR (AE)} & \num{3e-3} & 1500 & 5 \\
{LastFT-Reg (MMF)} & \num{1e-3} & 1000 & /  \\
{FDR (MMF)} & \num{1e-3} & 1000 & / \\
\bottomrule
\end{tabular}
\label{tab:hyper-utk}
\end{table}

\section{Additional Experiments}\label{sec:exp
_add}
\subsection{Assessing Fairness on UTKFace Dataset}
\label{sec:exp_utkface}
We also conduct the experiments on the UTKFace dataset with the goal of accurately predicting the age of the person in each image. The results of three different fairness notions are presented in Table~\ref{tab:utk}. Based on our experiments, we find that (1) FDR has the best fairness metrics and the best overall performance (indicated by the highest AF), across all three fairness metrics. This suggests that the proposed method is efficient in improving fairness of the deep neural network. (2) FDR and {LastFT-Reg} have relatively low test WACC compared to other methods, indicating that adding fairness constraints may have a negative impact on the model's prediction accuracy. 
\begin{table}[htb]
\caption{Last layer fine-tuning results with fairness notions on UTKFace dataset. Mean are reported over twenty random trials. Notably, methods including {LastFT} and {LastFT-RW} have the same value for WACC and AUC across different fairness notions since they do not depend on them, while other methods have different scores for different fairness notions.}       

\centering
\setlength{\tabcolsep}{4pt}
\begin{tabularx}{\linewidth}{l|CC|CC|CC|C}
\toprule
\textbf{Fairness Notion 1: EO} & \multicolumn{2}{c|}{WACC} & \multicolumn{2}{c|}{AUC} & \multicolumn{2}{c|}{EO\_Diff} & AF \\
\midrule
& Train & Test & Train & Test & Train & Test & Test \\
\midrule
FullFT-Reg & 0.998 & \textbf{0.804} & 1.000 & \textbf{0.892} & 0.002 &  0.141 & 0.663 \\
\midrule
{LastFT} & 0.821 & 0.793 & 0.902 & 0.880 & 0.126 & 0.152 &  0.641 \\
{LastFT-RW} & 0.864 & 0.797 & 0.938 & 0.878  & 0.069 & 0.104 & 0.693 \\
{LastFT-Reg} & 0.821 & 0.793 & 0.901 & 0.879 & 0.140 & 0.140 & 0.653 \\
FDR & 0.848 & 0.781 & 0.928 & 0.863 & 0.011 & \textbf{0.026} & \textbf{0.755} \\
\midrule
\midrule
\textbf{Fairness Notion 2: AE} & \multicolumn{2}{c|}{WACC} & \multicolumn{2}{c|}{AUC} & \multicolumn{2}{c|}{AE\_Diff} & AF \\
\midrule
& Train & Test & Train & Test & Train & Test & Test \\
\midrule
FullFT-Reg &  0.998 & \textbf{0.805} & 1.000 & \textbf{0.892} & 0.001 & 0.037 & 0.768 \\
\midrule
{LastFT} & 0.821 & 0.793 & 0.902 & 0.880  &  0.027 & 0.029 &  0.764 \\
{LastFT-RW} & 0.864 & 0.797 & 0.938 & 0.878 & 0.012 & 0.026 & 0.771 \\
{LastFT-Reg} &  0.822 & 0.791 & 0.901 & 0.878 & 0.010 & 0.028 & 0.763 \\
FDR & 0.857 & 0.785 & 0.936 & 0.866 & 0.003 & \textbf{0.011} & \textbf{0.774} \\
\midrule
\midrule
\textbf{Fairness Notion 3: MMF} & \multicolumn{2}{c|}{WACC} & \multicolumn{2}{c|}{AUC} & \multicolumn{2}{c|}{WA} & AF \\
\midrule
& Train & Test & Train & Test & Train & Test & Test \\
\midrule
FullFT-Reg & 0.999 & \textbf{0.814} & 1.000 & \textbf{0.899} & 0.997 & 0.710 & 1.524 \\
\midrule
{LastFT} & 0.821 & 0.793 & 0.902 & 0.880  & 0.736 & 0.689 & 1.482\\
{LastFT-RW} & 0.864 & 0.797 & 0.938 & 0.878 & 0.825 & 0.727 & 1.524 \\
{LastFT-Reg} & 0.807 & 0.779 & 0.892 & 0.866 & 0.776 & 0.726 & 1.505 \\
FDR & 0.851  & 0.788 & 0.930 & 0.870 & 0.829 & \textbf{0.745} & \textbf{1.533} \\
\bottomrule
\end{tabularx}
\label{tab:utk}
\end{table}
\subsection{Assessing Fairness under Transfer Learning Setting}~\label{trans}
To show that fairness can still be satisfied despite a change in data distribution, we explore the task that involves adapting a pre-trained model on the large ImageNet dataset to the CelebA dataset for the purpose of hair-color prediction, and the UTKFace dataset for age prediction, using last layer fine-tuning. We select Equalization Odds (EO) to illustrate our idea. The results are presented in Table~\ref{tab:trans_celeb}\&\ref{tab:trans_utk}. To avoid bias, the FullFT-Reg method is excluded from this task. From the results, Equalized Odds can be effectively maintained during OOD-fine-tuning using FDR in both two datasets. Additionally, FDR also achieves the best overall performance (highest AF value) without hurting the WACC and AUC significantly. 

\begin{table*}[htb]
\centering
\caption{Last layer fine-tuning results with fairness notions in the transfer learning setting on \textbf{CelebA} dataset. For WACC, AUC and AF, a larger value is considered better; while for EO\_Diff, a smaller value is considered better. Mean are reported over twenty random trials.}
\begin{tabularx}{\linewidth}{l|CC|CC|CC|C}
       \toprule
    \multicolumn{8}{l}{\textbf{Fairness Notion: Equalized Odds} (smaller value is better)} \\
    \midrule
       & \multicolumn{2}{c|}{ WACC } &  \multicolumn{2}{c|}{ AUC }  & \multicolumn{2}{c|}{ EO\_Diff } & AF \\ \midrule
    & Train & Test & Train & Test & Train & Test & Test \\ \midrule
    {LastFT} & 0.899 & \textbf{0.871} & 0.961 & \textbf{0.944} & 0.252 & 0.412 & 0.459     \\
    {LastFT-RW} & 0.871 & 0.856 & 0.943 & 0.930 & 0.070 & 0.114  &  0.742    \\
    {LastFT-Reg} & 0.893 &0.867  & 0.959 &0.941  & 0.168 &0.302   &    0.565    \\
    FDR & 0.854 & 0.841 & 0.925 &  0.915& 0.021 & \textbf{0.062}  &  \textbf{0.779} \\
    \bottomrule
\end{tabularx}
\label{tab:trans_celeb}
\end{table*}

\begin{table*}[htb]
\centering
\caption{Last layer fine-tuning results with fairness notions in the transfer learning setting on \textbf{UTKFace} dataset. For WACC, AUC and AF, a larger value is considered better; while for EO\_Diff, a smaller value is considered better. Mean are reported over twenty random trials.}
\begin{tabularx}{\linewidth}{l|CC|CC|CC|C}
       \toprule
    \multicolumn{8}{l}{\textbf{Fairness Notion: Equalized Odds} (smaller value is better)} \\
    \midrule
       & \multicolumn{2}{c|}{ WACC } &  \multicolumn{2}{c|}{ AUC }  & \multicolumn{2}{c|}{ EO\_Diff } & AF \\ \midrule
    & Train & Test & Train & Test & Train & Test & Test \\ \midrule
    {LastFT} & 0.751 & 0.695 & 0.833 & 0.766 & 0.174 & 0.185 & 0.510 \\
{LastFT-RW} & 0.714 & \textbf{0.702} & 0.790 & \textbf{0.775} & 0.113 & 0.127 & 0.575\\
{LastFT-Reg} & 0.729 & 0.687 & 0.817 & 0.761 & 0.214 & 0.197 & 0.490 \\
FDR & 0.695 & 0.682  & 0.768 &  0.752 & 0.011 &  \textbf{0.033} & \textbf{0.649} \\
    \bottomrule
\end{tabularx}
\label{tab:trans_utk}
\end{table*}

\subsection{Further Exploration with Surgical Fine-tuning }\label{sec:surg}

\citet{lee2022surgical} proposed a modification to the last layer fine-tuning by extending it to different blocks which consists of a set of layers of the model. For example, the ResNet-18 architecture can be divided into six blocks: an input layer which is an initial convolutional layer; Blocks 1, 2, 3 and 4, each of which comprised of multiple convolutional layers with batch normalization and activation functions, followed by a shortcut connection; and the last layer. During the fine-tuning, they select one block to update and fix the parameters of other layers. They also propose several criteria for determining the appropriate subset of layers to perform fine-tuning, such as Auto-RGN~\citep{lee2022surgical} which automatically selects an appropriate subset of layers for fine-tuning.

Inspired by that, we conduct the similar experiments in the fairness setting. In our experiments (results are shown in Table~\ref{tab:4}), different blocks of the model are fine-tuned on the balanced sampled dataset while incorporating fairness constraints. According to the results, fine-tuning any set of layers other than the input layer can effectively address the fairness over-fitting issue without affecting the ACC/AUC performance of the model. Fine-tuning Block 1 typically performs better than fine-tuning other blocks. This finding suggests that it is promising to explore more sophisticated fine-tuning strategies, which opens an interesting direction for future work. 
\begin{table*}[tbh]
\centering
        \caption{Surgical fine-tuning results with Equalized Odds fairness notion on balanced-sampled CelebA dataset. Mean are reported over twenty random trials.}
        \begin{tabularx}{\linewidth}{l|CCC|CCC|CCC}
        \toprule
       Metrics & \multicolumn{3}{c|}{ WACC } &  \multicolumn{3}{c|}{ AUC }  &  \multicolumn{3}{c}{ Fairness Metric}  \\ \midrule
    Fairness Constraint & EO & AE & MMF & EO & AE & MMF & EO & AE & MMF \\ \midrule
        Last Layer &  0.863 &  0.877  &  0.882 &  0.935  &  0.963  & 0.962  & 0.109   & 0.009  &   0.781    \\         \midrule
        Input Layer & 0.916 &  0.919  &  0.918 & 0.948 &  0.947 &  0.947 & 0.182   & 0.062  &  0.423   \\
        Block 1 & 0.903 & 0.906  &  0.886 & 0.963 & 0.963  & 0.962  & 
 \textbf{0.071} &  0.010  &   \textbf{0.860}  \\
        Block 2 & 0.896  &  0.907 &  0.890 &  0.965 & 0.964   &  0.964 &  0.081   &  0.011 &   0.833    \\
        Block 3 & 0.883  & 0.903  &  0.895 &  0.959 & 0.966   & 0.964  &  0.084    &  0.006 &    0.812    \\ 
        Block 4 &  0.883  & 0.890  &  0.895 &  0.957  & 0.961  & 0.961  &  0.135    & 0.010  &   0.770  \\
        Auto-RGN & 0.888  & 0.895  &  0.895 & 0.963 & 0.966 & 0.965 &  0.096    &  \textbf{0.005} & 0.778    \\
        \bottomrule
        \end{tabularx}
        \label{tab:4}

\end{table*}


\end{document}